\documentclass[letterpaper, 10 pt, journal, twoside]{IEEEtran}  
\usepackage[noadjust]{cite}

\newcommand{\blue}[1]{\textcolor{black}{#1}}
\usepackage{graphicx} 
\usepackage{subfigure}
\usepackage{amsmath} 
\usepackage{amssymb}  
\usepackage{amsthm}
\usepackage{multirow}
\usepackage{upgreek}
\usepackage{mathtools,lipsum, nccmath}
\usepackage[normalem]{ulem}
\usepackage{xcolor}
\usepackage{kotex}
\usepackage{pifont}

\usepackage{mathrsfs}
\usepackage{setspace}
\usepackage{afterpage }
\usepackage{balance}
\usepackage{float}

\newcommand{\todocite}[1]{\textcolor{red}{[TODO(cite)]}}

\title{\LARGE \bf
Geometric Tracking Control of Omnidirectional Multirotors for Aggressive Maneuvers}

\author{Hyungyu Lee, Sheng Cheng, Zhuohuan Wu, Jaeyoung Lim, Roland Siegwart, and Naira Hovakimyan
\thanks{This work was supported in part by NASA ULI under Grant 80NSSC22M0070, in part by NSF-AoF RI under Grant 2133656, and in part by NSF M3X under Grant 2431216. (Corresponding author: Sheng Cheng.)}
\thanks{H. Lee, S. Cheng, Z. Wu, and N. Hovakimyan are with the Department of Mechanical Science and Engineering, University of Illinois Urbana-Champaign, USA.
({\tt\small email: \{hyungyu2, chengs, zw24, nhovakim\}@illinois.edu})}
\thanks{J. Lim and R. Siegwart are with the Autonomous Systems Laboratory, ETH Zürich, Switzerland.
({\tt\small email: \{jalim, rolandsi\}@ethz.ch})}
\thanks{Video demonstration: https://youtu.be/0rpY1bxjegc}
}

\begin{document}

\maketitle
\begin{abstract}
An omnidirectional multirotor has the maneuverability of decoupled translational and rotational motions, superseding the traditional multirotors' motion capability. Such maneuverability is achieved due to the ability of the omnidirectional multirotor to frequently alter the thrust amplitude and direction. In doing so, the rotors' settling time, which is induced by inherent rotor dynamics, significantly affects the omnidirectional multirotor's tracking performance, especially in aggressive flights. To resolve this issue, we propose a novel tracking controller that takes the rotor dynamics into account and does not require additional rotor state measurement. \blue{This is achieved by integrating a linear rotor dynamics model into the vehicle's equations of motion and designing a PD controller to compensate for the effects introduced by rotor dynamics.} We prove that the proposed controller yields almost global exponential stability. The proposed controller is validated in experiments, where we demonstrate significantly improved tracking performance in multiple aggressive maneuvers compared with a baseline geometric PD controller.
\end{abstract}


\section{Introduction} \label{int}
Multirotor vehicles, also referred to as multirotors, are becoming widely used technologies in real-world applications for their simple mechanical structure, agility, and low cost.
As multirotors are brought to new application domains, there is a rising demand to further extend their maneuverability~\cite{caros,chanyoung,zhuohuan,kumar}. To fulfill this need, fully-actuated multirotors, \blue{which are vehicles capable of independent control over all six degrees of freedom (DoFs), three translational DoFs and three rotational DoFs, have been considered. These systems use} fixed-tilt~\cite{ryll3, othex2} or variable-tilt~\cite{ryll2, zhe} rotor systems that enable the vehicle to carry out translational motions without altering the attitude. While improving maneuverability, these platforms do not show significant attitude-changing capability, such as tilting over 30 degrees during hovering\blue{~\cite{zhe}}. To address this issue, omnidirectional multirotors that can generate thrust to cancel out their gravity at any attitude are gaining more attention~\cite{tog2}. We summarize recent research on omnidirectional multirotors in Table~\ref{table1}. The domain can be categorized as omnidirectional multirotors with unidirectional rotors~\cite{tog,uni7, taelee6,vol,vol2} or bidirectional rotors~\cite{stick,stick2,bres1,bres2,bresE}. 

The unidirectional rotors have been employed extensively since they are more accessible and have a higher power efficiency than the bidirectional ones.
Furthermore, they do not experience force exertion at low speeds, which results in reversing delay. Despite these advantages, the unidirectional rotors render the system mechanically complicated since omnidirectional flights require either at least seven fixed-tilt rotors or additional servo motors paired with each rotor to enable a variable-tilt rotor system~\cite{bres2}. The extra mechanical parts result in increased weight of the system and challenges in control, which is undesirable for multirotors. 
An alternative approach to address these challenges is to use bidirectional rotors, which keep the tilting fixed and allow the rotors to rotate in both directions, providing thrust in either direction along the rotation axis.
The bidirectional rotors offer an excellent solution to mitigate the mechanical complexity by unidirectional rotors \cite{stick,stick2,bres1,bres2,bresE}: 
No additional hardware is required to facilitate direction change or thrust aid. However, \blue{bidirectional rotors are less power efficient than unidirectional rotors}, and they suffer from the reversing delay~\cite{tog, bres2} (occurring when rotor rotation reverses).

\begin{table*}[ht!]
\centering
\footnotesize
TABLE \uppercase\expandafter{\romannumeral1\::}
\textsc{A Survey of Recent Work in Omnidirectional Multirotor Controls. The Abbreviations Quat., Rot., \blue{G.E.S and L.A.S.} Stand for Quaternion, Rotation Matrix, \blue{global exponential stability, and local asymptotic stability}, Respectively.}
    \label{table1}
     \vspace{0.2cm}
\small     
\begin{tabular}{r|llllll}
\hline\hline 
Method & Rotor-tilt Type & Propeller Type & Rotor Dynamics & Control Strategy & Stability Guarantee\\
\hline 
    \cite{tog} & Fixed-tilt & Unidirectional & \hspace{0.9cm}N & Geometric PID control with rot. & - \\
    O7+ \cite{uni7} & Fixed-tilt & Unidirectional &\hspace{0.9cm}N & Geometric PID control with rot. & - \\
    \cite{taelee6} & Fixed-tilt & Variable-pitch &\hspace{0.9cm}N & Geometric PID control with rot. & Almost G.E.S. \\
    Voliro \cite{vol} & Variable-tilt & Unidirectional &\hspace{0.9cm}N & Nonlinear PID control with quat. & - \\
    \cite{vol2} & Variable-tilt & Unidirectional &\hspace{0.9cm}N & LQR with integral action & L.A.S. \\
    ODAR-6 \cite{stick} & Fixed-tilt & Bidirectional &\hspace{0.9cm}N & Geometric PID control with rot. & - \\
    ODAR-8 \cite{stick2} & Fixed-tilt & Bidirectional &\hspace{0.9cm}N & Geometric PID control with rot. & - \\
    \cite{bres1,bresE} & Fixed-tilt & Bidirectional &\hspace{0.9cm}N & Nonlinear PID control with quat. & - \\
    \blue{\cite{ryll15novel}} & Variable-tilt & Unidirectional &\hspace{0.9cm}Y & \blue{Geometric PD control with rot.} & - \\
    \blue{\cite{li24servo}} & Variable-tilt & Unidirectional &\hspace{0.9cm}Y & \blue{Nonlinear MPC with quat.} & - \\
    \cite{bres2} & Fixed-tilt & Bidirectional &\hspace{0.9cm}Y & Nonlinear PID control with quat. & - \\
    \hline
    \textbf{Ours} & Fixed-tilt & Uni/Bidirectional &\hspace{0.9cm}Y & Geometric PD control with rot. & Almost G.E.S. \\ 
\hline\hline
\end{tabular}
\end{table*}

In this paper, we focus on performing aggressive maneuvers with an omnidirectional multirotor, which are essential for improving their operational capabilities in dynamic environments~\cite{muzza}.
The omnidirectional multirotors' aggressive motion requires the rotors (and actuators for variable-tilt rotors) to frequently and precisely change speed and even direction. 
The rotor and actuator dynamics, become significant and introduce disturbances to the vehicle's dynamics, which can lead to substantial tracking errors or even instability. 
These challenges call for the need to consider the rotor/actuator dynamics in motion control to achieve precise tracking of omnidirectional multirotors in aggressive flights.

\blue{To mitigate the effects of rotor/actuator dynamics, a common approach is to incorporate the dynamics into the motion controller design, as has been studied for fixed-wing aircraft~\cite{steffensen2023nonlinear} and quadrotors~\cite{muzza,torrente2021data}.
For omnidirectional multirotors with variable-tilt mechanisms, the servo actuators (for rotor tilting) have a slower response than rotors. Previous work~\cite{ryll15novel} applied a Smith predictor (as an effective tool~\cite{wolovich1993automatic} for handling known delays in a known stable linear system) to address servo delays.
Another study~\cite{li24servo} modeled the servo motor angle as a first-order system and employed nonlinear model predictive control (MPC) to mitigate these dynamics. 
In terms of fixed-tilt omnidirectional multirotors, rotor dynamics will be considered. 
In~\cite{bicego20nonlinear}, the authors developed a nonlinear MPC framework incorporating rotor dynamics by modeling the rotor angular velocity as a first-order system for generic multirotors. Similarly, for fixed-tilt omnidirectional multirotors with bidirectional rotors~\cite{bres2}, authors 
took a similar model
and implemented angular velocity feedback control. Both approaches demonstrate remarkable improvement. However, neither of them has shown a stability guarantee and experimental validation with an aggressive trajectory. Furthermore, their approaches need rotor state measurements, which require dedicated sensors or special Electronic Speed Controllers (ESCs).}

We propose a novel control architecture for omnidirectional multirotors, incorporating rotor dynamics directly into the controller design without requiring additional rotor state measurements.
We use a simplified rotor dynamics model and integrate it with the vehicle's equations of motion. 
Based on the combined equations of motion, we design the controller to accommodate the influence induced by the rotor dynamics. 
We prove that our architecture yields almost global exponential stability for the omnidirectional multirotor, unlike existing approaches that use a simplifying assumption that the rotors have fast (and thus negligible) dynamics~\cite{taelee6,vol2}.
We validate the proposed method in experiments, where it 
dramatically improves the tracking performance compared with a baseline controller (which neglects the rotor dynamics) for aggressive translational and rotational motions.
\blue{Our work is closely related to \cite{bres2}, where the authors present a comprehensive approach that incorporates rotor dynamics through a first-order DC motor model and implements an angular-velocity feedback control law, while minimizing the occurrence of rotor direction reversals. While their approach relies on rotor state feedback, our work demonstrates that stability in the presence of rotor dynamics can be achieved without these measurements, resulting in a simpler yet effective control framework.}

Our contributions are summarized as follows. i) We present the first omnidirectional multirotor controller design that accounts for rotor dynamics without requiring additional rotor state measurements. Furthermore, we prove the almost global exponential stability of the proposed controller with the complete system that includes the rotor dynamics. ii) We validate the proposed controller's performance using an eight-rotor omnidirectional multirotor in experiments.

\section{Modeling}\label{mod}
In this section, we provide the vehicle's dynamical model, including the rigid-body dynamics, the rotor dynamics, and the propeller's aerodynamics. To simplify the modeling, we assume the platform is rigid; The desired trajectory is smooth and differentiable; The rotors do not affect other rotors and do not saturate. These assumptions are widely used in the fully-actuated multirotor studies \cite{othex2,vol,vol2}. \blue{We also assume that reversing delay is negligible during aggressive flights, as we conclude from our preliminary experiments that rotor settling time is a major factor rather than the reversing delay.}

\begin{figure}[t!] 
    \centering
        {\hspace{0.55cm}
            \includegraphics[scale=0.45]{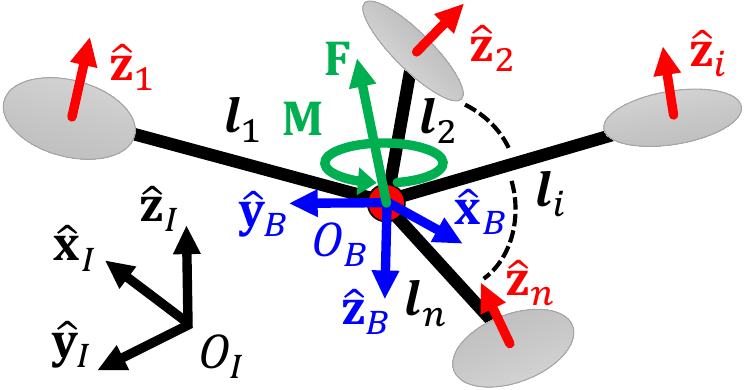}
        }
        \caption{Coordinate frames of a general omnidirectional multirotor, where $n$ denotes the number of rotors and $\boldsymbol{l}_i \in \mathbb{R}^{3}$ is the position of rotor $i$ from $O_B$. Note that $n\geq 6$ should hold for bidirectional rotors and $n\geq 7$ for unidirectional ones.}
        \label{coord} 
    \end{figure}

\blue{The thrust and torque generated by each rotor are commonly modeled as follows \cite{kumar,TaeLee}:}

\begin{align} 
    \label{eq3}
        f_i    
        =
        \mu \operatorname{sgn}(\Omega_i) {\Omega_i}^2,
        \quad 
        \tau_i
        =
        \kappa \operatorname{sgn}(\Omega_i) {\Omega_i}^2,
\end{align}
where $\mu \in \mathbb{R}$ and $\kappa\in \mathbb{R}$ denote the lift and drag coefficients of the rotor, respectively; $f_i\in \mathbb{R}$ and $\tau_i\in \mathbb{R}$ are the thrust force and the torque generated by the $i$-th rotor, respectively; $\Omega_i$ is the angular speed of rotor $i$; and $\operatorname{sgn}(\cdot)$ denotes the sign function. Note that \blue{for a unidirectional rotor, where $\Omega_i \geq{0}$ (i.e., the rotor spins in one direction), \eqref{eq3} reduces to $f_i = \mu {\Omega_i}^2$ and $\tau_i = \kappa{\Omega_i}^2$. For a bidirectional rotor, $\Omega_i \in \mathbb{R}$, as the rotor can spin in both directions.} The positive direction of $\Omega_i$ aligns with the $\mathbf{z}$-axis of the $i$-th rotor as shown in~Fig.~\ref{coord}.

\begin{figure}[t!] 
    \centering 
        \includegraphics[scale=0.53]{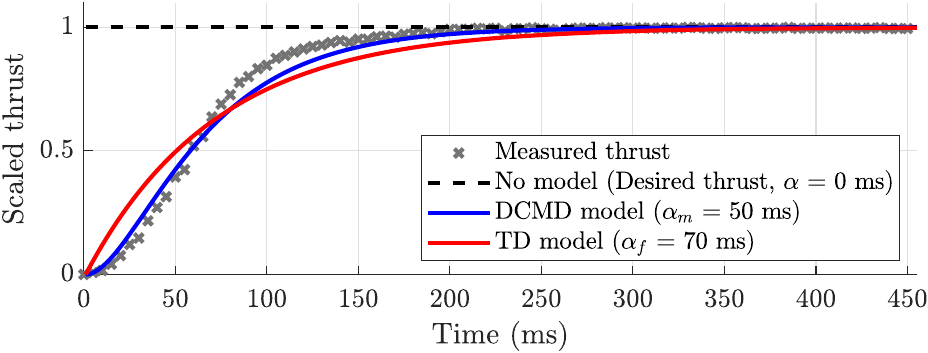} 
        \caption{
        Step response of a rotor with a propeller. The thrust from the rotor is scaled by the maximum thrust generated from the rotor. A 2000KV brushless rotor attached with Gemfan 513D 3-blade 3D propeller has been used for the measurement. Note that $\alpha$ represents the rotor time constant, and the subscripts $f$ and $m$ indicate the time constants of the TD and DCMD models, respectively. The desired and the no-model thrusts are the same, as the latter does not consider rotor dynamics. The time constants are determined experimentally.}   
        \label{response} 
     \end{figure}

There are several approaches to modeling the dynamics of a rotor~\cite{dcmotor,muzza}. A commonly used approach is modeling the rotor angular speed as a first-order system using brushless DC-motor dynamics~\cite{dcmotor}, which we refer to as the DCMD model. \blue{Using DCMD model, the rotor dynamics can be written as follows:}

\begin{equation}
    \label{dcmd}
\blue{\dot{\Omega}_{i}=\frac{1}{\alpha_{m,i}}\left(\Omega_{{cmd}, i}-{\Omega_{i}}\right),}
\end{equation}
\blue{where $\Omega_{{cmd},i}\in \mathbb{R}$ is commanded thrust, and $\alpha_{\blue{m,}i} \in \mathbb{R^+}$ is the rotor time constant for the $i$-th rotor.} The DCMD model \blue{accurately captures} the measured thrust \blue{characteristics}, as shown in Fig.~\ref{response}. However, due to the nonlinear relationship between the thrust and rotor speed, using the DCMD model-based method introduces complexity in the controller's structure, ultimately resulting in a less practical controller.

Another approach, referred to as the thrust dynamics (TD) model, simplifies the model by treating the thrust as a first-order system \cite{muzza}. The results therein show that satisfactory flight performance is preserved, although the TD model is a simplified model to ease controller design. Accordingly, as shown in Fig.1, although the DCMD model has the best performance, the TD model performs competently with no excessive errors compared to the conventional model-free method. Hence, we apply the TD model and write it as:   
\begin{equation}
    \label{fi}
\dot{f}_{i}=\dfrac{1}{\alpha_{\blue{f,}i}}\left(f_{{cmd}, i}-f_{i}\right),
\end{equation}
where $f_{{cmd},i}\in \mathbb{R}$ is commanded thrust, and $\alpha_{\blue{f,}i} \in \mathbb{R^+}$ is the thrust time constant for the $i$-th rotor.

Figure~\ref{coord} shows the coordinate frame of a generalized fixed-tilt omnidirectional platform. We define the global fixed frame or the inertial frame $\mathscr{F}_{I}=O_I,\{\hat{\mathbf{x}}_I,\hat{\mathbf{y}}_I,\hat{\mathbf{z}}_I\}$, and the body frame $\mathscr{F}_{B}=O_B,\{\hat{\mathbf{x}}_B,\hat{\mathbf{y}}_B,\hat{\mathbf{z}}_B\}$, where $O_B$ is located at the center of mass (CoM) of the omnidirectional multirotor. We also define $\mathscr{F}_{i}=O_i,\{\hat{\mathbf{x}}_i,\hat{\mathbf{y}}_i,\hat{\mathbf{z}}_i\}$ as the $i$-th rotor's frame expressed in $\mathscr{F}_{B}$.

Under the rigid-body assumption, the Newton-Euler formulation can be written as follows:

\begin{align} 
    \begin{split}
        \label{pR}
         \blue{\dot{\mathbf{p}}}&\:\blue{=\dot{\mathbf{v}},} \quad \blue{\dot{\mathbf{R}}=\mathbf{R} [\boldsymbol{\omega}]^\wedge,} 
    \end{split}\\
    \begin{split}
        \label{eom}
         m\dot{\mathbf{v}}&=-mg\hat{\mathbf{z}}_I+\mathbf{R}\mathbf{F},
    \end{split}\\
    \begin{split}
        \label{eom2}
         \mathbf{J}\dot{\boldsymbol{\omega}}&=-\boldsymbol{\omega} \times \mathbf{J} \boldsymbol{\omega}+\mathbf{M},
    \end{split}
\end{align}
where $\mathbf{p}=[x, y, z]^T \in \mathbb{R}^3$ and $\mathbf{v}=[v_x, v_y, v_z]^T$ $\in \mathbb{R}^3$ are the position and the linear velocity of the vehicle's CoM in the inertial frame, $\mathbf{R}\in \text{SO}(3)$ is the rotation matrix from frame $\mathscr{F}_{B}$ to $\mathscr{F}_{I}$, $m \in \mathbb{R^+}$ and $\mathbf{J}\in \mathbb{R}^{3\times3}$ are the mass and inertial tensor of the platform, respectively, $\boldsymbol{\omega}=[\omega_x, \omega_y, \omega_z]^T \in \mathbb{R}^3$ is the angular velocity, and $\mathbf{F}=[F_x, F_y, F_z]^T\in \mathbb{R}^3$ and $\mathbf{M}=[M_x, M_y, M_z]^T\in \mathbb{R}^3$ are the force and moment applied at CoM expressed in \blue{the body frame}, respectively. Note that $[\cdot]^{\wedge}:\mathbb{R}^{3} \rightarrow \mathfrak{so}(3)$ is the \textit{wedge operator} that maps a vector into a skew-symmetric matrix.

The relationship between the applied force $\mathbf{F}$ and moment $\mathbf{M}$ on the vehicle's body and rotor thrusts $\{f_i\}^n_{i=1}$ is: 

\begin{align}
    \begin{split}
    \label{eq8}
        \mathbf{F}=\sum_{i=1}^{n}f_i\hat{\mathbf{z}}_i,   
        \quad \mathbf{M}= \sum_{i=1}^{n}(\boldsymbol{l}_i \times f_i\hat{\mathbf{z}}_i+\tau_i\hat{\mathbf{z}}_i),
    \end{split}
\end{align}

\noindent 
which can be expressed in the matrix form as $[\mathbf{F}, \mathbf{M}]^T =\textbf{A}\boldsymbol{f}$, where $\boldsymbol{f}={[f_1, f_2, \cdots, f_n]}^T$ and $\mathbf{A}\in{\mathbb{R}}^{6 \times n}$ is the allocation matrix. Using the approximation that $\alpha_i\approx\alpha$, the single thrust dynamics model (\ref{fi}) can be expanded to collective wrench dynamics as follows:

\begin{equation}\label{wrench}
\dot{\mathbf{w}}=\frac{1}{\alpha}\left(\mathbf{w}_{{cmd}}-\mathbf{w}\right),
\end{equation}

\noindent
where $\mathbf{w}=[\mathbf{F}, \mathbf{M}]^T$, $\mathbf{w}_{cmd}=[\mathbf{F}_{cmd}, \mathbf{M}_{cmd}]^T$, and $\mathbf{F}_{cmd}$ and $\mathbf{M}_{cmd}$ are the commanded force and moment, respectively. With (\ref{eom}) and (\ref{wrench}), the vehicle's equations of motion with rotor dynamics can be established as follows:

\begin{align}
    \begin{split}
    \label{eomF}
         m \dot{\mathbf{v}}=&{} - \alpha \mathbf{R}\dot{\mathbf{F}} - mg\hat{\mathbf{z}}_I + \mathbf{R}\mathbf{F}_{cmd},
    \end{split}\\
    \begin{split}
    \label{eomM}
         \mathbf{J}\dot{\boldsymbol{\omega}}
         =&{}
         -\alpha \dot{\mathbf{M}} - \boldsymbol{\omega} \times \mathbf{J} \boldsymbol{\omega} +  \mathbf{M}_{cmd}.
    \end{split}
\end{align}
By including rotor dynamics in the Newton-Euler equations~\eqref{eom} and~\eqref{eom2}, $- \alpha \mathbf{R}\dot{\mathbf{F}}$ and $-\alpha \dot{\mathbf{M}}$ appear in the equation of motion, which deteriorate the tracking performance.
Note that $\dot{\mathbf{F}}$ and $\dot{\mathbf{M}}$ can be further broken down by differentiating and rearranging~\eqref{eom} and~\eqref{eom2} as follows:

\begin{align}
    \dot{\mathbf{F}} &= \mathbf{R}^T\left(m \ddot{\mathbf{v}} - \dot{\mathbf{R}} \mathbf{F}\right) = \mathbf{R}^T m \ddot{\mathbf{v}} + \mathbf{F} \times \boldsymbol{\omega}, \label{dotF}\\
    \mathbf{\dot{M}} &=\dot{ \boldsymbol{\omega}} \times \mathbf{J}\boldsymbol{\omega} + \boldsymbol{\omega} \times \mathbf{J}\dot{\boldsymbol{\omega}} + \mathbf{J}\ddot{\boldsymbol{\omega}}. \label{dotM}
\end{align}

Using the results of~\eqref{dotF} and~\eqref{dotM}, the complete equations of motion, considering the effect of rotor dynamics, can be written as follows:

\begin{align}
   m\dot{\mathbf{v}} &= -\alpha m \ddot{\mathbf{v}} - \alpha \mathbf{R}(\mathbf{F} \times \boldsymbol{\omega}) - mg\hat{\mathbf{z}}_I + \mathbf{R} \mathbf{F}_{\text{cmd}},\label{eomFF}\\
  \mathbf{J}\dot{\boldsymbol{\omega}} &=
         -\alpha \left(\dot{\boldsymbol{\omega}} \times \mathbf{J}\boldsymbol{\omega} + \boldsymbol{\omega} \times \mathbf{J}\dot{\boldsymbol{\omega}} + \mathbf{J}\ddot{\boldsymbol{\omega}}  \right) - \boldsymbol{\omega} \times \mathbf{J} \boldsymbol{\omega} +  \mathbf{M}_{cmd}. \label{eomMM}
\end{align}
From \eqref{eomFF} and \eqref{eomMM}, we can conclude that the adverse effects of rotor dynamics are particularly pronounced when there are large linear and angular jerks ($\ddot{\mathbf{v}}$ and $\ddot{\boldsymbol{\omega}}$), significant misalignment between the force and angular velocity vectors (i.e., $\mathbf{F} \times \boldsymbol{\omega}$), substantial changes in the direction of the angular velocity vector ($\dot{\boldsymbol{\omega}}$), and a large rotor time constant ($\alpha$), as shown in~\eqref{eomFF} and~\eqref{eomMM}. Such conditions are prevalent during aggressive flight maneuvers, leading to further degradation of tracking performance if rotor dynamics are neglected. To mitigate these issues, we incorporate compensatory measures in the controller design, as detailed in Section~\ref{geo}.

\section{Geometric Tracking Control} \label{geo}
In this section, we provide a control method for the omnidirectional multirotor to track the desired pose based on the modeling from Section~\ref{mod}. Unlike conventional multirotor controllers~\cite{TaeLee, ACRL}, tracking commands are carried out in two independent control loops since the translational and rotational dynamics are decoupled. We used a geometric PD controller, where we do not apply the integral term as it can amplify the rotor's settling time due to its integral nature. Furthermore, we define force and moment errors to take rotor dynamics into account in the controller design. Due to space limitations, we provide a more detailed derivation of the controller and the stability analysis in~\cite{arXiv2}.

\begin{figure}[t!]
      \centering
      \includegraphics[scale=0.3]{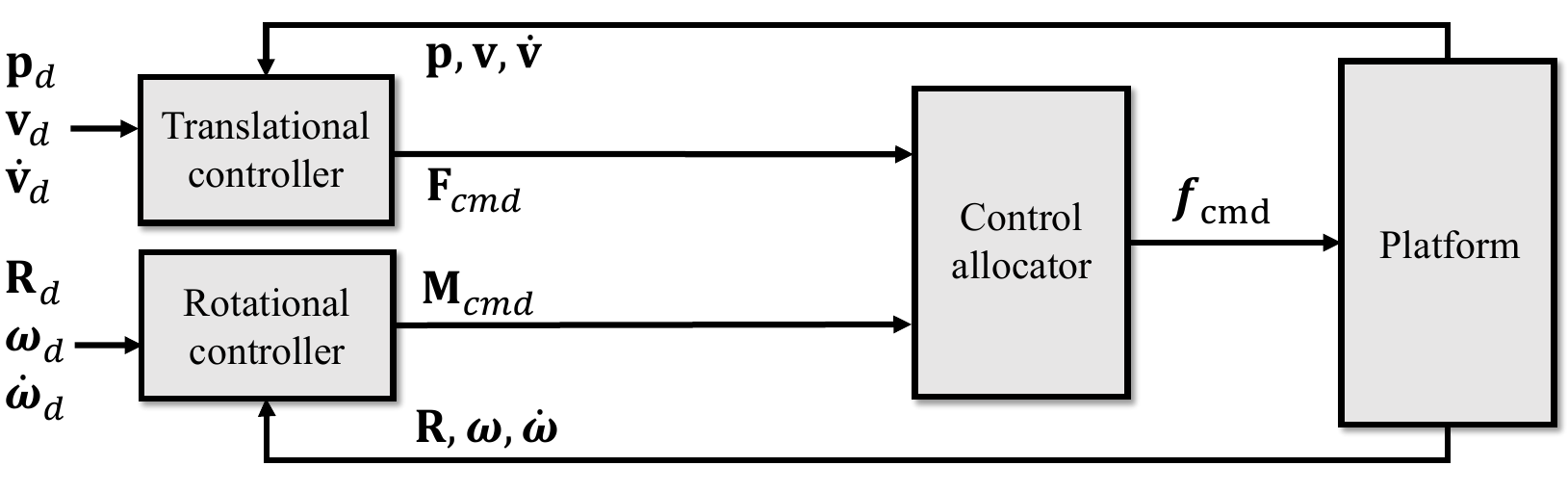}
      \caption{The control diagram of the proposed controller.}
      \label{diagram}
   \end{figure}
   
Figure~\ref{diagram} shows the overall control diagram of the proposed controller. The inputs $\mathbf{p}_{d}=[x_d,y_d,z_d]^T \in \mathbb{R}^3$, $\mathbf{v}_{d}=[v_{x_d},v_{y_d},v_{z_d}]^T \in \mathbb{R}^3$, and $\dot{\mathbf{v}}_{d}$ are fed to the position controller along with the actual position $\mathbf{p}$ and velocity $\mathbf{v}$. Simultaneously, the inputs $\mathbf{R}_d\in \text{SO}(3)$, $\boldsymbol{\omega}_d =[\omega_{x_d},\omega_{y_d},\omega_{z_d}]^T \in \mathbb{R}^3$, and $\dot{\boldsymbol{\omega}}_d \in \mathbb{R}^3$ are fed to the attitude controller along with actual rotation matrix $\mathbf{R}$ and angular velocity $\boldsymbol{\omega}$. Outputs from position and attitude controllers, commanded force $\mathbf{F}_{\textrm{cmd}}$ and moment $\mathbf{M}_{\textrm{cmd}}$, are sent to the control allocator to generate thrusts on all rotors $\boldsymbol{f}_{cmd}$ by $\boldsymbol{f}_{cmd}=\mathbf{A}^{\dagger}[\mathbf{F}_{\blue{{cmd}}}^T,\mathbf{M}_{\blue{cmd}}^T]^T$, where $\mathbf{A}^{\dagger}$ is pseudo-inverse  of allocation matrix $\mathbf{A}$.

\subsection{Translational Controller}
For the translational controller, we define the position error $\mathbf{e}_{p}=[e_x,e_y,e_z]^T\in \mathbb{R}^3$, velocity error $\mathbf{e}_{v}=[e_{v_x},e_{v_y},e_{v_z}]^T\in \mathbb{R}^3$, and force error $\mathbf{e}_{F} \in \mathbb{R}^3$ by $\mathbf{e}_{p}=\mathbf{p}-\mathbf{p}_{d}$, $\mathbf{e}_{v}=\mathbf{v}-\mathbf{v}_{d}$, and $\mathbf{e}_{F}=\mathbf{F}-\mathbf{F}_{d}$, where the desired force $\mathbf{F}_{d} \in \mathbb{R}^3$ is defined as
\begin{equation}
    \label{Fd}
\mathbf{F}_{ {d}}=\mathbf{R}^T\left(-{k}_{p} \mathbf{e}_{p}-{k}_{v} \mathbf{e}_{v}+m g \hat{\mathbf{z}}_I+m \dot{\mathbf{v}}_{d}\right),
\end{equation}
\noindent
for $k_p\in \mathbb{R^+}$ and $k_v\in \mathbb{R^+}$ being position and velocity gains, respectively. 

To accommodate the settling time of the generated force, the force controller is designed as follows:

\begin{equation}
    \label{Fcmd}
\mathbf{F}_{{cmd}}=\mathbf{F}_{{d}}+\alpha \dot{\mathbf{F}}_{{d}},
\end{equation}
\noindent
where $\dot{\mathbf{F}}_d$, via differentiating~\eqref{Fd}, can be expressed as: 

\begin{align}
\label{Fddot}
    \dot{\mathbf{F}}_d =&\: \dot{\mathbf{R}}^T\left(-{k}_{p} \mathbf{e}_{p}-{k}_{v} \mathbf{e}_{v}+m g \hat{\mathbf{z}}_I+m \dot{\mathbf{v}}_{d}\right) \nonumber \\
    &+\mathbf{R}^T\frac{d}{dt} \left(-{k}_{p} \mathbf{e}_{p}-{k}_{v} \mathbf{e}_{v}+m g \hat{\mathbf{z}}_I+m \dot{\mathbf{v}}_{d}\right), \nonumber\\
    =&\: -[\boldsymbol{\omega}]^{\wedge}\mathbf{F}_d + \mathbf{R}^T \left( -k_p \mathbf{e}_v - \frac{k_v}{m} \mathbf{e}_F + m \ddot{\mathbf{v}}_d \right),\\
    =&\: \mathbf{F}_d \times \boldsymbol{\omega} + \mathbf{R}^T \left( -k_p \mathbf{e}_v - \frac{k_v}{m} \mathbf{e}_F + m 
    \ddot{\mathbf{v}}_d \right).
\end{align}

Applying~\eqref{Fddot} to~\eqref{Fcmd}, the complete form of control law can be written as: 

\begin{align}
\mathbf{F}_{\text{cmd}} &= \mathbf{R}^T \bigg( -k_p \mathbf{e}_p - (\alpha k_p + k_v) \mathbf{e}_v - \frac{\alpha k_v}{m} \mathbf{e}_F \nonumber \\
   &\quad + mg \mathbf{z}^I + m \dot{\mathbf{v}}_d + \alpha \mathbf{R} (\mathbf{F}_d \times \boldsymbol{\omega}) + \alpha m \ddot{\mathbf{v}}_d \bigg).
\end{align}

Note that $\ddot{\mathbf{v}}_{{d}}$ can be analytically computed from the given trajectory as the desired trajectory is 3rd-order and continuous. Considering that $\mathbf{e}_{F}$, which can be estimated using onboard IMU, is subject to sensor noise that is uncorrelated with trajectory tracking, filtering methods (such as a first-order low-pass filter) can be applied to suppress high-frequency noise. The specific parameters of the filter can be adjusted based on the application's requirements and the characteristics of the sensor noise.

\subsection{Rotational Controller}
For the rotational controller, we define the attitude error $\mathbf{e}_{R}=[e_{R_x},e_{R_y},e_{R_z}]^T\in \mathbb{R}^3$, angular velocity error $\mathbf{e}_{\omega}=[e_{\omega_x},e_{\omega_y},e_{\omega_z}]^T\in \mathbb{R}^3$, and moment error $\mathbf{e}_{M} \in \mathbb{R}^3$ as $
\mathbf{e}_{R} = $ \small$\frac{1}{2}\left[\mathbf{R}_{d}^{T} \mathbf{R}-\mathbf{R}^{T} \mathbf{R}_{d}\right]^{\vee}/\sqrt{1+\text{tr}(\mathbf{R}_{d}^{T} \mathbf{R})}$, \normalsize
$ \mathbf{e}_{\omega}=\boldsymbol{\omega}-\mathbf{R}^{T} \mathbf{R}_{d} \boldsymbol{\omega}_{d}$, and
$ \mathbf{e}_{M}=\mathbf{M}-\mathbf{M}_{d}$. The \textit{vee operator} $[\cdot]^{\vee}:\mathfrak{s o}(3) \rightarrow \mathbb{R}^{3}$ is the inverse of the \textit{wedge operator} and $\operatorname{tr}(\cdot)$ is the trace of a square matrix. The error $\mathbf{e}_{R}$ was first proposed in~\cite{LargeAtt}, and it is mathematically equivalent to the quaternion-based attitude error~\cite{arXiv}. We define the desired moment $\mathbf{M}_{{d}}$ as

\begin{multline}
    \label{Md}
\mathbf{M}_{d}= -k_{R} \mathbf{e}_{R}-k_{\omega} \mathbf{e}_{\omega}+{\boldsymbol{\omega}} \times \mathbf{J} {\boldsymbol{\omega}}\\
-\mathbf{J}\left(\left[{\boldsymbol{\omega}}\right]^{\wedge} \mathbf{R}^{T} \mathbf{R}_{d} {\boldsymbol{\omega}}_{d}-\mathbf{R}^{T} \mathbf{R}_{d} \dot{{\boldsymbol{\omega}}}_{d}\right),
\end{multline}
\noindent
where $k_R\in \mathbb{R^+}$ and $k_\omega\in \mathbb{R^+}$ are attitude and angular velocity gains, respectively.

To resolve the settling time of the generated moment, the moment controller is designed as follows:

\begin{equation}
    \label{Mcmd}
\mathbf{M}_{ {cmd}}=\mathbf{M}_{{d}}+\alpha \dot{\mathbf{M}}_{{d}}.
\end{equation}

We use the following equalities in our derivation:
\begin{align}
&\frac{d}{d t}\left(\mathbf{R}^T \mathbf{R}_d\right) = \mathbf{R}^T 
\mathbf{R}_d [\boldsymbol{\omega}_d]^{\wedge}-[\boldsymbol{\omega}]^{\wedge} \mathbf{R}^T \mathbf{R}_d, \label{RtRd}\\
&\frac{d}{d t}\left(\mathbf{R}^T \mathbf{R}_d \boldsymbol{\omega}_d\right) 
=\mathbf{R}^T \mathbf{R}_d \dot{\boldsymbol{\omega}}_d-[\boldsymbol{\omega}]^{\wedge} \mathbf{R}^T \mathbf{R}_d \boldsymbol{\omega}_d,\label{RtRdwd}\\
&\frac{d}{d t}\left(\mathbf{R}^T \mathbf{R}_d \nonumber
\dot{\boldsymbol{w}}_d\right) =\frac{d}{d t}\left(\mathbf{R}^T \mathbf{R}_d\right) \dot{\boldsymbol{w}}_d+\mathbf{R}^T \mathbf{R}_d\ddot{\boldsymbol{w}}_d\\
 &=\mathbf{R}^T \mathbf{R}_d [\boldsymbol{\omega}]^{\wedge}_d \dot{\boldsymbol{w}}_d-[\boldsymbol{\omega}]^{\wedge} \mathbf{R}^T \mathbf{R}_d \dot{\boldsymbol{w}}_d+\mathbf{R}^T \mathbf{R}_d \ddot{\boldsymbol{w}}_d \\
&\frac{d}{d t}\left([\boldsymbol{\omega}]^{\wedge} \mathbf{R}^T \mathbf{R}_d\nonumber \boldsymbol{w}_d\right)\\ 
&=[\dot{\boldsymbol{\omega}}]^{\wedge} \mathbf{R}^T \mathbf{R}_d \boldsymbol{w}_d+[\boldsymbol{\omega}]^{\wedge} \frac{d}{d t}\left(\mathbf{R}^T \mathbf{R}_d \boldsymbol{w}_d\right)\nonumber\\
& =[\boldsymbol{\omega}]^{\wedge} \mathbf{R}^T \mathbf{R}_d \boldsymbol{w}_d+[\boldsymbol{\omega}]^{\wedge} \mathbf{R}^T \mathbf{R}_d \dot{\boldsymbol{w}}_d-[\boldsymbol{\omega}]^{\wedge} [\boldsymbol{\omega}]^{\wedge} \mathbf{R}^T \mathbf{R}_d \boldsymbol{w}_d.
\end{align}

By differentiating~\eqref{Md}, $\dot{\mathbf{M}}_d$ can be expressed as follows: 

\begin{align}
    \label{Mddot}
        \dot{\mathbf{M}}_{d} =& -k_{R} \dot{\mathbf{e}}_{R}-k_{\omega} \dot{\mathbf{e}}_{\omega}+{\boldsymbol{\omega}} \times \mathbf{J} \dot{\boldsymbol{\omega}} + \dot{{\boldsymbol{\omega}}} \times \mathbf{J}\boldsymbol{\omega}\nonumber\\
&+\mathbf{J} \big( \left[{\boldsymbol{\omega}}\right]^{\wedge}\left[{\boldsymbol{\omega}}\right]^{\wedge} \mathbf{R}^T \mathbf{R}_d \boldsymbol{\omega}_d - \left[{\dot{\boldsymbol{\omega}}}\right]^{\wedge} \mathbf{R}^T \mathbf{R}_d \boldsymbol{\omega}_d \nonumber\\
&\:\mathbf{R}^T \mathbf{R}_d \left[{\boldsymbol{\omega}_d}\right]^{\wedge}\dot{\boldsymbol{\omega}}_d
- 2\left[{\boldsymbol{\omega}}\right]^{\wedge} \mathbf{R}^T \mathbf{R}_d \dot{\boldsymbol{\omega}}_d + \mathbf{R}^T \mathbf{R}_d \ddot{\boldsymbol{\omega}}_d \big),
\end{align}

\noindent
where, based on~\cite{LargeAtt}, $\dot{\mathbf{e}}_{R}$ and  $\dot{\mathbf{e}}_{\omega}$ are: 

\begin{align}
\dot{\mathbf{e}}_{R} &= \frac{\left(\operatorname{tr}\left[\mathbf{R}^T \mathbf{R}_d\right] \mathbf{I} - \mathbf{R}^T \mathbf{R}_d + 2 \mathbf{e}_{R} \mathbf{e}_{R}^T\right)}{2 \sqrt{1+\operatorname{tr}\left[\mathbf{R}_d^T \mathbf{R}\right]}} \mathbf{e}_{\omega},\\
\dot{\mathbf{e}}_{\omega} &= \dot{\boldsymbol{\omega}} + [\boldsymbol{\omega}]^{\wedge} \mathbf{R}^T \mathbf{R}_d \boldsymbol{\omega}_d - \mathbf{R}^T \mathbf{R}_d \dot{\boldsymbol{\omega}}_d.
\end{align}

  Note that $\mathbf{e}_{M}$ can be estimated using the derivative of vehicle angular velocity. Similar to $\mathbf{e}_{F}$, $\mathbf{e}_{M}$ is subject to sensor noise that is uncorrelated with trajectory tracking, and hence filtering methods (such as a first-order low-pass filter) can be applied to suppress high-frequency noise. 

\subsection{Stability Analysis}

For the stability analysis, we first analyze the stability of the translational and rotational control systems individually and then combine the results for the full system's stability.

\textit{Proposition 1:} (Global exponential stability of the translational system) Consider the commanded force $\mathbf{F}_{{cmd}}$ defined in (\ref{Fcmd}). If the positive design constants $k_p$, $k_v$, and $c_1$ satisfy 

\begin{align}
    \label{kp}
k_{p}>\frac{c_{1} k_{v}^{2}+2 c_{1}k_{v}-c_{1}^{2}}{m\left(4\left(k_{v}-c_{1}\right)-1\right)},\quad k_{v}>c_1+\frac{1}{4},
\end{align}

\noindent
then the zero equilibrium of the translational tracking error dynamics of $\mathbf{e}_{p}, \mathbf{e}_{v}$, and $\mathbf{e}_F$ is globally exponentially stable.

\noindent
\begin{proof}Let a Lyapunov function candidate $\mathcal{V}_{1}$ for the translational system be 
\begin{equation}
\mathcal{V}_{1}=\frac{1}{2} k_{p}\left\|\mathbf{e}_{p}\right\|^{2}+\frac{1}{2} m\left\|\mathbf{e}_{v}\right\|^{2}+\frac{1}{2} \alpha\left\|\mathbf{e}_{F}\right\|^{2}+ c_{1}  \mathbf{e}_{p} \cdot \mathbf{e}_{v},
\end{equation} 
where $\left\|\cdot\right\|$ denotes the Euclidean norm.

By Cauchy-Schwartz inequality, we can show that, for $k_p$ satisfying (\ref{kp}), $\mathcal{V}_{1}$ is bounded by

\begin{equation}
    \label{v1}
\mathbf{z}_{1}^{T} \mathbf{M}_{11} \mathbf{z}_{1} \leq \mathcal{V}_{1} \leq \mathbf{z}_{1}^{T} \mathbf{M}_{12} \mathbf{z}_{1},
\end{equation}

\noindent
where $\mathbf{z}_{1}=\left[\left\|\mathbf{e}_{p}\right\|,\left\|\mathbf{e}_{v}\right\|,\left\|\mathbf{e}_{F}\right\|\right]^{T} \in \mathbb{R}^{3}$, and the matrices $\mathbf{M}_{1 1} \text{ and } \mathbf{M}_{1 2} \in \mathbb{R}^{3 \times 3}$ are defined as
$$
\mathbf{M}_{11}=\frac{1}{2}\begin{bmatrix}
k_{p} & \hspace{-0.05 cm}-c_{1} & 0 \\
\hspace{-0.05 cm}-c_{1} & m & 0 \\
 0 & 0 & \alpha 
\end{bmatrix}, \quad
\mathbf{M}_{12}=\frac{1}{2}\begin{bmatrix}
k_{p} & c_{1} & 0 \\
c_{1} & m & 0 \\
 0 & 0 & \alpha
\end{bmatrix}. 
$$

Now we deal with the boundedness of $\dot{\mathcal{V}}_{1}$. From (\ref{wrench}), (\ref{Fd}), and (\ref{Fcmd}), the force error dynamics can be written as

\begin{equation}
\begin{aligned}
\label{eF'}
\alpha \dot{\mathbf{e}}_{F}&=\alpha\dot{\mathbf{F}}-\alpha\dot{\mathbf{F}}_{d}=({\mathbf{F}}_{cmd}-{\mathbf{F}})-\alpha\dot{\mathbf{F}}_{d}\\
&=(\mathbf{F}_{d}+\alpha\dot{\mathbf{F}}_{d}-\mathbf{F})-\alpha\dot{\mathbf{F}}_{d}=-\left(\mathbf{F}-\mathbf{F}_d\right) \\
&= -\mathbf{e}_{F}.
\end{aligned}
\end{equation}
\noindent
By (\ref{eomF}), (\ref{Fd}), (\ref{Fcmd}), and (\ref{eF'}), the velocity error dynamics can be written as

\begin{equation}
\begin{aligned}
\label{ev'}
m \dot{\mathbf{e}}_{v}&=m\dot{\mathbf{v}}-m\dot{\mathbf{v}}_d =-\alpha \mathbf{R}\dot{\mathbf{F}} - mg\hat{\mathbf{z}}_I + \mathbf{R}\mathbf{F}_{cmd} -m\dot{\mathbf{v}}_d \\
&=-k_{p} \mathbf{e}_{p}-k_{v} \mathbf{e}_{v}-\alpha\mathbf{R} \left(\dot{\mathbf{F}}-\dot{\mathbf{F}}_{d}\right)\\
&=-k_{p} \mathbf{e}_{p}-k_{v} \mathbf{e}_{v}+\mathbf{R} \mathbf{e}_{F}.
\end{aligned}
\end{equation}

\noindent
Using (\ref{eF'})--(\ref{ev'}), the time derivative of $\mathcal{V}_1$ is given by

\begin{equation}
\begin{aligned}
\dot{\mathcal{V}}_1 =&\: k_{p} \mathbf{e}_{p} \cdot \mathbf{e}_{v}+m \mathbf{e}_{v} \cdot \dot{\mathbf{e}}_{v}+\alpha \mathbf{e}_{F} \cdot \dot{\mathbf{e}}_{F}\\
&+c_{1}\left(\dot{\mathbf{e}}_{p} \cdot \mathbf{e}_{v}+\mathbf{e}_{p} \cdot \dot{\mathbf{e}}_{v}\right) \\
=&\:k_{p} \mathbf{e}_{p} \cdot \mathbf{e}_{v}+\mathbf{e}_{v} \cdot\left(-k_{p} \mathbf{e}_{p}-k_{v} \mathbf{e}_{v}+\mathbf{R} \mathbf{e}_{F}\right)  -\mathbf{e}_{F} \cdot \mathbf{e}_{F}\\
&+c_{1}\left(\dot{\mathbf{e}}_{p} \cdot \mathbf{e}_{v}+\mathbf{e}_{p} \cdot \frac{1}{m}\left(-k_{p} \mathbf{e}_{p}-k_{v} \mathbf{e}_{v}+\mathbf{R} \mathbf{e}_{F}\right)\right) \\
=&-\frac{c_{1} k_{p}}{m}\left\|\mathbf{e}_{p}\right\|^{2}-\left(k_{v}-c_{1}\right)\left\|\mathbf{e}_{v}\right\|^{2} -\left\|\mathbf{e}_{F}\right\|^{2}\\
&+\frac{c_{1} k_{v}}{m} \mathbf{e}_{p} \cdot \mathbf{e}_{v}+\mathbf{R} \mathbf{e}_{F} \cdot\left(\mathbf{e}_{v}+\frac{c_{1}}{m} \mathbf{e}_{p}\right).
\end{aligned}
\end{equation}
Since $\left\|\mathbf{R} \mathbf{e}_{F}\right\| = \left\|\mathbf{e}_{F}\right\|$ holds by the property of rotation matrix~\cite{rot}, we can show that $\dot{\mathcal{V}}_{1}$ is bounded by

\begin{equation}
\begin{aligned}
\label{v1'}
\dot{\mathcal{V}}_{1} \leq -\mathbf{z}_{1}^{T} \mathbf{W}_{1} \mathbf{z}_{1},
\end{aligned}
\end{equation}

\noindent
where
$$
\mathbf{W}_{1}=\begin{bmatrix}
\vspace{0.1 cm}
c_{1} k_{p} / m & \hspace{-0.15 cm}-c_{1} k_{v} / (2 m) & \hspace{-0.15 cm}-c_{1} / (2 m) \\
\vspace{0.1 cm}
\hspace{-0.05 cm}-c_{1} k_{v} / (2 m) & k_{v}-c_{1} & \hspace{-0.15 cm}-1 / 2 \\
\hspace{-0.05 cm}-c_{1} / (2 m) & \hspace{-0.15 cm}-1 / 2 & 1
\end{bmatrix}.
$$

\noindent 
With the positive design constants $k_p$, $k_v$, and $c_1$ satisfying~\eqref{kp}, the matrices $\mathbf{M}_{11}, \mathbf{M}_{12}, \text{and } \mathbf{W}_{1}$ are always positive definite. As a result, $\mathcal{V}_{1}$ is always bounded as in~(\ref{v1}), and $\dot{\mathcal{V}}_{1}$ is always negative within the region of attraction $\mathbf{z}_{1} \in \mathbb{R}^3$. Thus, the translational motion of the system is globally exponentially stable~\cite{stable}.
\end{proof}

To show the stability of the rotational system, we first define the rotational error function between two rotation matrices, $\mathbf{R}_1$ and $\mathbf{R}_2$, as follows:

\begin{equation}
\Psi\left(\mathbf{R}_1, \mathbf{R}_2\right)=2-\sqrt{1+\text{tr}(\mathbf{R}_{d}^{T} \mathbf{R})}.
\end{equation}

Note that $\Psi\left(\mathbf{R}_1, \mathbf{R}_2\right)$ is bounded by $0 \leq \Psi\left(\mathbf{R}_1, \mathbf{R}_2\right) \leq 2$, and $\Psi\left(\mathbf{R}_1, \mathbf{R}_2\right) = 2$ if and only if the minimum angle required to rotate from $\mathbf{R}_1$ to $\mathbf{R}_2$ is 180 degrees.

\textit{Proposition 2:} (Almost global exponential stability of the rotational system) Consider the commanded moment $\mathbf{M}_{{cmd}}$ defined in (\ref{Mcmd}). If the vehicle's initial rotation matrix $\mathbf{R}$ satisfies 

\begin{equation}
\label{psy}
\Psi\left(\mathbf{R}(0), \mathbf{R}_{d}(0)\right)<2,
\end{equation}

\noindent
and the positive design constants $k_R$, $k_\omega$, and $c_2$ satisfy

\begin{align}
\label{kr}
k_{R}>\frac{c_2 k_{\omega}^2}{\lambda_{m}\left(4\left(k_{\omega}-\frac{1}{2}c_{2}\right)-1\right)}, \quad k_{\omega}>\frac{1}{2}c_{2}+\frac{1}{4},
\end{align}

\noindent
where $\lambda_{m}$ denotes the minimum eigenvalue of the inertia tensor~$\mathbf{J}$, then the zero equilibrium of the rotational tracking error dynamics of $\mathbf{e}_{R}, \mathbf{e}_{\omega}$, and $\mathbf{e}_M$ is almost globally exponentially stable. 

\begin{proof}Let a Lyapunov function candidate for the rotational system $\mathcal{V}_{2}$ be 
\begin{equation}
\mathcal{V}_{2}=\frac{1}{2} \mathbf{e}_{\omega} \cdot \mathbf{J} \mathbf{e}_{\omega}+k_{R} \Psi\left(\mathbf{R}, \mathbf{R}_{d}\right)+\frac{1}{2}\alpha\left\|\mathbf{e}_{M}\right\|^2+c_{2} \mathbf{e}_{R} \cdot \mathbf{e}_{\omega}.
\end{equation}

\noindent
In~\cite{LargeAtt}, it has been shown that $ \Psi\left(\mathbf{R}, \mathbf{R}_{d}\right)$ is bounded by 

\begin{equation}
\label{ePsy}
\left\|\mathbf{e}_{R}\right\|^{2} \leq \Psi\left(\mathbf{R}, \mathbf{R}_{d}\right) \leq 2\left\|\mathbf{e}_{R}\right\|^{2}.
\end{equation}

Using (\ref{ePsy}) and $\lambda_{{m}}\left\|\mathbf{e}_{\omega}\right\|^{2} \leq \mathbf{e}_{\omega} \cdot \mathbf{J} \mathbf{e}_{\omega} \leq \lambda_{{M}}\left\|\mathbf{e}_{\omega}\right\|^{2}$, where $\lambda_{M}$ is the maximum eigenvalue of $\mathbf{J}$, we can show that, for $k_R$ and $k_\omega$ satisfying (\ref{kr}), $\mathcal{V}_{2}$ is bounded by

\begin{equation}
\label{v2}
\mathbf{z}_{2}^{T} \mathbf{M}_{21} \mathbf{z}_{2} \leq \mathcal{V}_{2} \leq \mathbf{z}_{2}^{T} \mathbf{M}_{22} \mathbf{z}_{2},
\end{equation}

\noindent
where $\mathbf{z}_{2}=\left[\left\|\mathbf{e}_{R}\right\|,\left\|\mathbf{e}_{\omega}\right\|,\left\|\mathbf{e}_{M}\right\|\right]^{T} \in \mathbb{R}^{3}$, and the matrices $\mathbf{M}_{2 1} \text{ and } \mathbf{M}_{2 2} \in \mathbb{R}^{3 \times 3}$ are defined as

$$
\mathbf{M}_{21}=\frac{1}{2}\begin{bmatrix}
2k_{R} & \hspace{-0.05 cm}-c_{2} & 0 \\
\hspace{-0.05 cm}-c_{2} & \lambda_m & 0 \\
 0 & 0 & \alpha 
\end{bmatrix},
\mathbf{M}_{22}=\frac{1}{2}\begin{bmatrix}
4k_{R} & c_{2} & 0 \\
c_{2} & \lambda_M & 0 \\
 0 & 0 & \alpha
\end{bmatrix}. 
$$

Now we deal with the boundedness of $\dot{\mathcal{V}}_{2}$. In~\cite{LargeAtt}, it has been shown that the following relationships hold:

\begin{align}
\begin{split}
\label{ePsy'}
\dot{\Psi}\left(\mathbf{R}, \mathbf{R}_{d}\right)={}\mathbf{e}_{R} \cdot \mathbf{e}_{\omega}, \quad\left\|\dot{\mathbf{e}}_{R}\right\|  \leq \frac{1}{2}\left\|\mathbf{e}_{\omega}\right\|.
\end{split}
\end{align}

\noindent
From (\ref{wrench}), (\ref{Md}), and (\ref{Mcmd}), the moment error dynamics can be written as

\begin{equation}
\begin{aligned}
\label{eM'}
\alpha \dot{\mathbf{e}_{M}}&=\alpha\dot{\mathbf{M}}-\alpha\dot{\mathbf{M}_{d}}=({\mathbf{M}}_{cmd}-{\mathbf{M}})-\alpha\dot{\mathbf{M}}_{d}\\
&=(\mathbf{M}_{d}+\alpha\dot{\mathbf{M}}_{d}-\mathbf{M})-\alpha\dot{\mathbf{M}_{d}}=-(\mathbf{M}-\mathbf{M}_d)\\
&= -\mathbf{e}_{M}.
\end{aligned}
\end{equation}

\noindent
From (\ref{eomM}), (\ref{Md}), (\ref{Mcmd}), and (\ref{eM'}), the angular velocity error dynamics can be written as

\begin{equation}
\begin{aligned}
\label{ew'}
\mathbf{J} \dot{\mathbf{e}}_{\omega}&=\mathbf{J} \dot{\boldsymbol{\omega}}-\mathbf{J} \frac{d}{dt}\left(\mathbf{R}^{T} \mathbf{R}_{d} \boldsymbol{\omega}_{d}\right)\\
&=\mathbf{J} \dot{\boldsymbol{\omega}}+\mathbf{J}\left(\left[{\boldsymbol{\omega}}\right]^{\wedge} \mathbf{R}^{T} \mathbf{R}_{d} \boldsymbol{\omega}_{d}-\mathbf{R}^{T} \mathbf{R}_{d} \dot{\boldsymbol{\omega}}_{d}\right)\\
&=-k_{R} \mathbf{e}_{R}-k_{\omega} \mathbf{e}_{\omega}-\alpha\left(\dot{\mathbf{M}}-\dot{\mathbf{M}}_{d}\right)\\
&=-k_{R} \mathbf{e}_{R}-k_{\omega} \mathbf{e}_{\omega}+\mathbf{e}_{M}.
\end{aligned}
\end{equation}

\noindent
Using (\ref{ePsy'})--(\ref{ew'}), the time derivative of $\mathcal{V}_2$ is given by

\begin{equation}
\begin{aligned}
\dot{\mathcal{V}}_{2}=&\:\mathbf{e}_{\omega} \cdot\left(-k_{R} \mathbf{e}_{R}-k_{\omega} \mathbf{e}_{\omega}+\mathbf{e}_{M}\right)+k_{R} \mathbf{e}_{R} \cdot \mathbf{e}_{\omega}-\mathbf{e}_{M}\cdot \mathbf{e}_{M}\\
&+c_{2} \dot{\mathbf{e}}_{R} \cdot \mathbf{e}_{\omega}+c_{2}\mathbf{e}_{R}\cdot   \mathbf{J}^{-1}\left(-k_{R} \mathbf{e}_{R}-k_{\omega} \mathbf{e}_{\omega}+\mathbf{e}_{M}\right) \\
=&-k_{\omega}\left\|\mathbf{e}_{\omega}\right\|^{2}-\left\|\mathbf{e}_{M}\right\|^{2}-c_{2} k_{R} \mathbf{e}_{R} \cdot \mathbf{J}^{-1} \mathbf{e}_{R}+c_{2} \dot{\mathbf{e}}_{R} \cdot \mathbf{e}_{\omega}\\
&-c_{2} k_{\omega} \mathbf{e}_{R} \cdot \mathbf{J}^{-1} \mathbf{e}_{\omega}+\mathbf{e}_{M} \cdot \mathbf{e}_{\omega}+c_{2} \mathbf{e}_{R} \cdot \mathbf{J}^{-1} \mathbf{e}_{M}.
\end{aligned}
\end{equation}

\noindent
As $\left\|\dot{\mathbf{e}}_{R}\right\|\leq \frac{1}{2}\left\|\mathbf{e}_{\omega}\right\|$, we can show that $\dot{\mathcal{V}}_{2}$ is bounded by

\begin{equation}
\begin{aligned}
\label{v2'}
\dot{\mathcal{V}}_{2} \leq -\mathbf{z}_{2}^{T} \mathbf{W}_{2} \mathbf{z}_{2},
\end{aligned}
\end{equation}

\noindent
where
$$
\mathbf{W}_{2}=\begin{bmatrix}
\vspace{0.1 cm}
c_{2} k_{R} / \lambda_{m} & c_{2} k_{\omega} / (2 \lambda_{m}) & \hspace{-0.15 cm}-c_{2} / (2 \lambda_{M}) \\
\vspace{0.1 cm}
c_{2} k_{\omega} / (2 \lambda_{m}) & k_{\omega} - c_{2}/2  & \hspace{-0.15 cm}-1 / 2 \\
\hspace{-0.05 cm}-c_{2} / (2 \lambda_{M}) & \hspace{-0.15 cm}-1 / 2 & 1
\end{bmatrix}.
$$

With the positive design constants $k_R$, $k_\omega$, and $c_2$ satisfying (\ref{kr}), the matrices $\mathbf{M}_{21}, \mathbf{M}_{22}, \text{and } \mathbf{W}_{2}$ are always positive definite. As a result, $\mathcal{V}_{2}$ is always bounded as in~(\ref{v2}), and $\dot{\mathcal{V}}_{2}$ is always negative within the region of attraction in (\ref{psy}). Thus, the rotational motion of the system is almost globally exponentially stable.
\end{proof}

\textit{Theorem 1:} (Almost global exponential stability of the full system) Consider the commanded force $\mathbf{F}_{{cmd}}$ and the commanded moment $\mathbf{M}_{{cmd}}$ defined in (\ref{Fcmd}) and (\ref{Mcmd}). If positive design constants $c_1$, $c_2$, $k_p$, $k_v$, $k_R$, and $k_{\omega}$ satisfy (\ref{kp}) and (\ref{kr}), then the zero equilibrium of the tracking error dynamics of $\mathbf{e}_{p}$, $\mathbf{e}_{v}$, $\mathbf{e}_{R}$, $\mathbf{e}_{\omega}$, $\mathbf{e}_F$, and $\mathbf{e}_M$ is exponentially stable. 

\begin{proof} Let a Lyapunov function candidate $\mathcal{V}$ for the full system be 
\begin{equation}
\mathcal{V}= \mathcal{V}_{1}+\mathcal{V}_{2}.
\end{equation}

\noindent
Using (\ref{v1}) and (\ref{v2}), the function  $\mathcal{V}$ can be bounded

\begin{equation}
\label{v_all}
\mathbf{z}_{1}^{T} \mathbf{M}_{11} \mathbf{z}_{1} + \mathbf{z}_{2}^{T} \mathbf{M}_{21} \mathbf{z}_{2} \leq \mathcal{V} \leq \mathbf{z}_{1}^{T} \mathbf{M}_{12} \mathbf{z}_{1}+\mathbf{z}_{2}^{T} \mathbf{M}_{22} \mathbf{z}_{2}.
\end{equation}

\noindent
From (\ref{v1'}) and (\ref{v2'}), the time derivative of $\mathcal{V}$ is bounded by 

\begin{equation}
\begin{aligned}
\dot{\mathcal{V}} \leq -\mathbf{z}_{1}^{T} \mathbf{W}_{1} \mathbf{z}_{1} -\mathbf{z}_{2}^{T} \mathbf{W}_{2} \mathbf{z}_{2}.
\end{aligned}
\end{equation}

\noindent
With the positive design constants satisfying (\ref{kp}) and (\ref{kr}), the matrices $\mathbf{M}_{11}$, $\mathbf{M}_{12}$, $\mathbf{M}_{21}$, $\mathbf{M}_{22}$, $\mathbf{W}_{1}$, and $\mathbf{W}_{2}$ are always positive definite. Hence, $\mathcal{V}$ is always bounded as in~(\ref{v_all}), and $\dot{\mathcal{V}}$ is always negative within the region of attraction in (\ref{psy}). Thus, the full system is almost globally exponentially stable with the region of attraction in~(\ref{psy}).
\end{proof}

\section{Experimental Validation} \label{exp} 
In this section, we present the experimental results validating the proposed controller's performance. We compare the proposed controller with a baseline controller for tracking performance. The baseline controller {considers} the rotors {being} ideal with no settling time, resulting in using $\mathbf{F}_d$ and $\mathbf{M}_d$ as control inputs. For the proposed rotational controller, we used the simplified form of $\dot{\mathbf{M}}_d$. 
Note that by assuming that the rotational tracking control performance is sufficiently good so that $\mathbf{R}^T \mathbf{R}_d \approx \mathbf{I}$, $\dot{\mathbf{M}}_{{d}}$ can be simplified to: 

\begin{align}
        \dot{\mathbf{M}}_{d} &\approx -\frac{1}{2}k_{R}{\mathbf{e}}_{\omega}-k_{\omega} \mathbf{J}^{-1}{\mathbf{e}}_{M}+{\boldsymbol{\omega}} \times \mathbf{J} \dot{\boldsymbol{\omega}} +\dot{{\boldsymbol{\omega}}} \times \mathbf{J}\boldsymbol{\omega} +\mathbf{J}\ddot{\boldsymbol{\omega}}_d.
\end{align}

For the omnidirectional multirotor platform, we used the configuration with eight fixed-tilt bidirectional rotors proposed in~\cite{bres1}, as it has been shown to maximize the vehicle's agility while making its characteristics nearly rotationally invariant. For the platform's main electrical components, the system incorporates eight BrotherHobby LPD 2306.5 2000KV rotors paired with Gemfan 513D 3-blad 3D propellers, two Tekko32 F4 4in1 50A ESCs, and Pixhawk as a flight controller. Position data were captured using external Vicon cameras. Regarding hardware parameters, the rotor time constant is $\alpha_f=0.07$~s (as shown in Fig.~1). \blue{While these time constants can be relatively slow compared to smaller rotors, they represent the actual characteristics of our experimental setup.} \blue{The maximum rotor thrust is $f_\textit{{max}}=12.6$~N,} the mass is $m=1.481$~kg, the moment of inertia matrix is $\mathbf{J}=\operatorname{diag}(0.020, 0.021, 0.020)$~kg$\cdot$m$^2$, and the arm length $||\boldsymbol{l}{i}||=0.15$~m. For the control software, we utilized the PX4 Autopilot as a baseline. 

\blue{For all experiments, translational and rotational controllers are running at $100 \text{ Hz}$ and $800 \text{ Hz}$, respectively. We utilized a first-order low-pass filter running at 800 $\text{Hz}$} with a cutoff frequency of 40~\text{Hz} to remove high-frequency noise from $\mathbf{e}_{F}$ and $\mathbf{e}_{M}$\footnote{ 
    \blue{Note that $\mathbf{e}_{F}$ and $\mathbf{e}_{M}$ are majorly determined by the trajectory’s frequencies, which are relatively low, e.g., 0.5--1 Hz for trajectories in the following experiments. Given the low-frequency nature of $\mathbf{e}_{F}$ and $\mathbf{e}_{M}$, the implementation of a low-pass filter with a 40-Hz cutoff frequency can effectively suppress high-frequency sensor noise while introducing negligible phase delay for the control purpose.}}. The nominal control gains were set to $k_p = 10$, $k_v = 3.7$ for translational controller, and $k_R = 3.07$, $k_\omega = 0.315$ for rotational controller. Both the baseline and proposed controllers utilized identical control gains throughout all experiments. \blue{Note that our platform's ESCs do not provide rotor state measurements.
    }

The experimental validation consisted of three main test cases: a purely translational trajectory, a single-axis rotational trajectory, and a multi-axis rotational trajectory. These tests were designed to highlight different aspects of how rotor dynamics affect the multirotor's performance in position and attitude control. The last multi-axis rotational test showcased the combined effects on both position and attitude control during aggressive maneuvers.

\subsection{Tracking a Purely Translational Trajectory}\label{exp2}
\begin{figure}[t!]
    \centering
        \subfigure
        { \hspace{-0.1 cm}
            \includegraphics[scale=0.7]{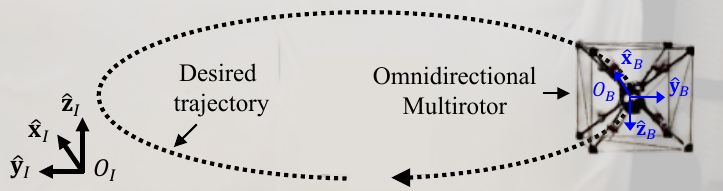}
        }\vspace{-0.45cm}
        \begin{minipage}{0.49\linewidth}
          \begin{figure}[H]
              \includegraphics[width=\linewidth]{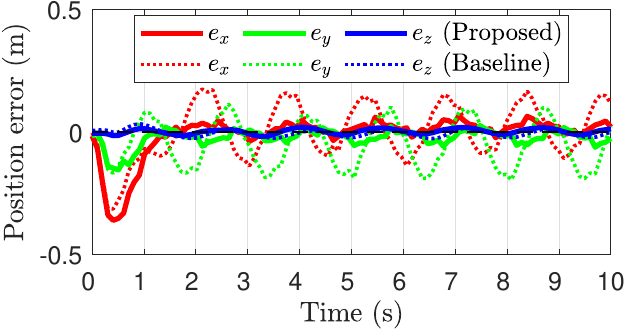}
              \end{figure}
      \end{minipage}\vspace{-0.1cm}
      \begin{minipage}{0.49\linewidth}
          \begin{figure}[H]
             \includegraphics[width=\linewidth]{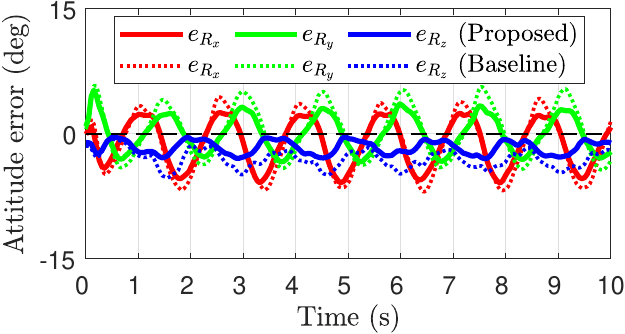}
              \end{figure}
      \end{minipage}  \vspace{-0.1cm}
        
    \caption{(L-R): Comparison of the omnidirectional multirotor’s position and attitude error between the proposed and baseline controllers over the purely translation trajectory shown on top. Position error of the proposed method reveals significantly lower transient and steady-state tracking errors in the $x$ and $y$ directions, highlighting the enhanced translational tracking performance. \blue{Attitude errors are shown via Euler angles for clear interpretation.}}
    \label{fig4}
\end{figure}

\blue{
This experiment is conducted to validate the translational control system under the influence of rotor dynamics. Initially, the platform started at $\mathbf{p}(0)=[-0.4,0,0.6]^T \text{ m}$, following the desired trajectory $x_d(t) = -0.4\:\text{cos}(\frac{4\pi}{3} t) \text{ m}$, $y_d(t) = 0.4\:\text{sin}(\frac{4\pi}{3} t) \text{ m}$ and $z_d(t) = 0.6 \text{ m}$. To exclude the impact of rotational dynamics on the translational system, we set the desired attitude $\mathbf{R}_d = \mathbf{I}$. This trajectory allows us to exclude rotational dynamics while examining the performance of the translational controller.}

Figure 4 presents the results of this experiment, illustrating the superior tracking performance of the proposed controller along the $\hat{\mathbf{x}}_I$ and $\hat{\mathbf{y}}_I$ axes. Unlike the baseline controller, which exhibits significant tracking errors in these axes due to its inability to counteract the effect of rotor dynamics, our controller demonstrates robust compensation capabilities, achieving \blue{31}\% less position root-mean-squared error (RMSE) compared to the baseline approach. As discussed in Section~\ref{mod}, these disturbances are exacerbated by longer rotor settling time (bigger $\alpha$) and are more pronounced with the translational jerk $\ddot{\mathbf{v}}$ specific to the given trajectory. Both controllers maintain consistent performance along the $\hat{\mathbf{z}}_I$-axis, where a fixed height of 0.6 meter renders the $\alpha m \ddot{\mathbf{v}}$ in~(\ref{eomFF}) negligible. Similarly, because the desired rotation matrix remains fixed, $\dot{\mathbf{M}}$ is zero, resulting in comparable rotational tracking performance for both controllers. These results demonstrate the enhanced translational tracking capability of the proposed controller and its ability to mitigate the adverse control effects due to rotor dynamics.

\subsection{Tracking a Single-axis Rotational Trajectory}

\begin{figure}[t!]
    \centering
         \subfigure
        { \hspace{-0.1 cm}
        
            \includegraphics[scale=0.7]{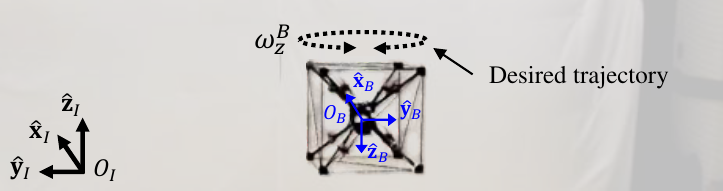}
        }\vspace{-0.45cm}
        \begin{minipage}{0.49\linewidth}
          \begin{figure}[H]
              \includegraphics[width=\linewidth]{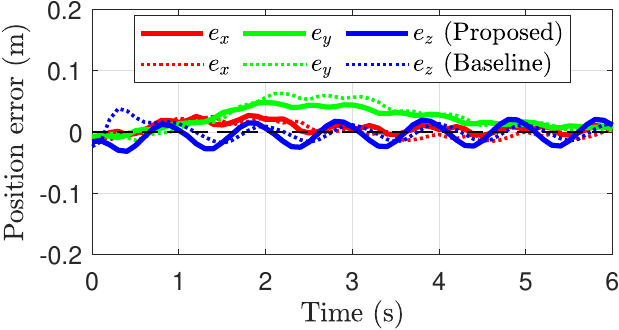}
              \end{figure}
      \end{minipage}\vspace{-0.1cm}
      \begin{minipage}{0.49\linewidth}
          \begin{figure}[H]
             \includegraphics[width=\linewidth]{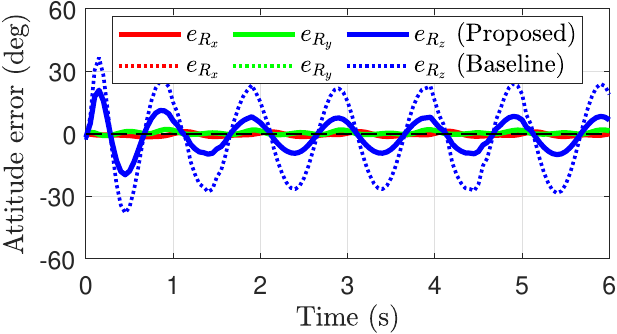}
              \end{figure}
      \end{minipage} 
     
    \caption{(L-R): The comparison of position and attitude tracking errors of the omnidirectional multirotor with the proposed and baseline controllers over the single-axis rotational trajectory shown on top. The proposed controller demonstrates superior attitude tracking performance, particularly around the z-axis where high angular jerk is applied while maintaining comparable positional stability with regards to the baseline controller.}
    \label{fig5}
\end{figure}

To validate the rotational control system under the influence of rotor dynamics, particularly focusing on the effects of a high angular jerk, we designed this experiment, which restricts the rotation to a single axis. The platform, initially hovering at position $\mathbf{p}(0)=[0,0,1]^T \text{m}$, followed a desired angular velocity $\omega^B_z = \frac{\pi}{2}(\sin(2\pi t))\, \text{rad/s}$, where the superscripts $B$ indicate the body frame. This trajectory was specifically chosen to generate significant angular jerk, signifying the term $\mathbf{J}\ddot{\boldsymbol{\omega}}$ in equation~\eqref{dotM}, while not exciting other terms induced from rotor dynamics. 

Figure~\ref{fig5} illustrates the results of this experiment. The impact of rotor dynamics was prominently observed in the attitude control, particularly around the $z_B$-axis where the high angular jerk was applied. The baseline controller exhibited significant attitude tracking errors due to its inability to account for the rotor dynamics in the presence of high angular acceleration and jerk. In contrast, the proposed controller demonstrated superior performance, substantially reducing the attitude RMSE by 39\%.
This marked improvement in attitude control can be attributed to the proposed controller's compensation for the effects of rotor dynamics, particularly the term $\mathbf{J}\ddot{\boldsymbol{\omega}}$ in \eqref{eomMM}. By incorporating these dynamics into the control design, the proposed method was able to anticipate and counteract the disturbance introduced by rotor dynamics, resulting in more precise attitude tracking during the aggressive rotational maneuver. Since the translational dynamics are not directly influenced by an angular jerk and the rotation axis is parallel to gravity resulting in $\mathbf{F} \times \boldsymbol{\omega} = 0$, both controllers maintained similar positional stability throughout the maneuver.

\subsection{Tracking a Multi-axis Rotational Trajectory}\label{exp4}
\begin{figure}[t]
    \centering 
         \subfigure 
        { \hspace{-0.0 cm}
            \includegraphics[scale=0.69]{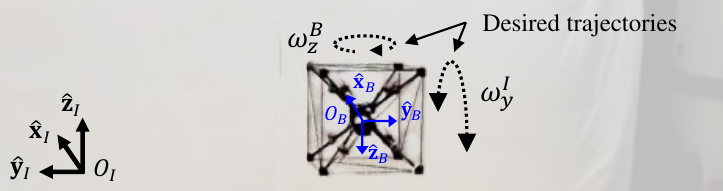}
        }\vspace{-0.45cm}
      \begin{minipage}{0.48\linewidth}
          \begin{figure}[H]
              \includegraphics[width=\linewidth]{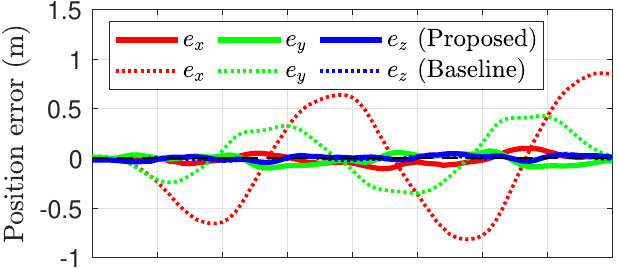}
              \end{figure}
      \end{minipage}\vspace{-0.15cm}
      \begin{minipage}{0.48\linewidth}
          \begin{figure}[H]
             \includegraphics[width=\linewidth]{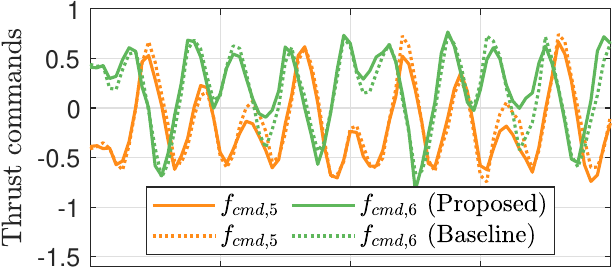}
              \end{figure}
      \end{minipage}\vspace{-0.15cm}
      \begin{minipage}{0.48\linewidth}
          \begin{figure}[H]
              \includegraphics[width=\linewidth]{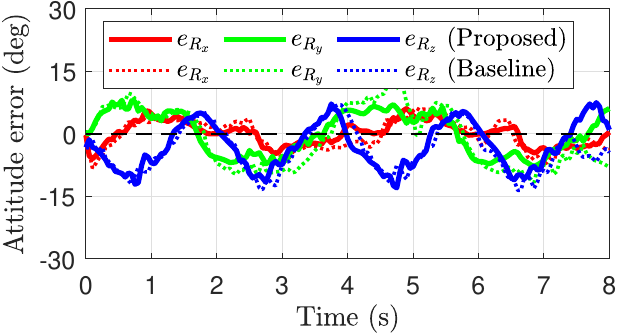}
              \end{figure}
      \end{minipage}
      \begin{minipage}{0.48\linewidth}
          \begin{figure}[H]
             \includegraphics[width=\linewidth]{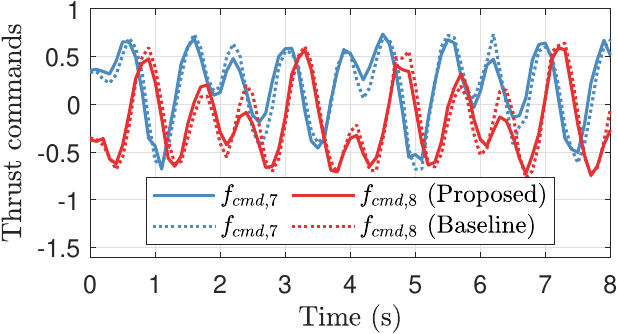}
              \end{figure}
      \end{minipage}
     
    \caption{\blue{Comparison between the proposed and baseline controllers showing (Left) position and attitude tracking errors and (Right) commanded thrusts for rotors 5-8 during the multi-axis rotational trajectory depicted on top.} The proposed controller's position and attitude errors highlight the improved tracking performance of the proposed method. \blue{The commanded thrust comparison (with the magnitude of 1 representing maximum thrust) reveals that the proposed controller generates thrust commands with phase lead, compared with the baseline, for compensating for the rotor settling time.}} 
    \label{fig6} 
\end{figure}

We designed a trajectory to signify terms in both translational and rotational dynamics that are impacted by the rotor dynamics, including $\mathbf{F} \times \boldsymbol{\omega}$, $\dot{\boldsymbol{\omega}} \times \mathbf{J}\boldsymbol{\omega} + \boldsymbol{\omega} \times \mathbf{J}\dot{\boldsymbol{\omega}}$, and $\mathbf{J}\ddot{\boldsymbol{\omega}}$ shown in~\eqref{eomFF} and~\eqref{eomMM}. The vehicle followed a desired angular velocity $\omega^B_z = 2\pi \, \text{rad/s}$ and $\omega^I_y(t) = \frac{\pi}{2} \sin(\pi t) \, \text{rad/s}$, where the superscripts $I$ indicate that the vector components are expressed in the inertial frames. 

Figure~\ref{fig6} illustrates the results. The baseline controller's position errors are due to rotor settling times. As the platform rotates, delayed rotor response redirects $m g \hat{\mathbf{z}}_I$ that was intended for gravity compensation. Consequently, this misalignment results in undesirable forces along the $\hat{\mathbf{x}}_I$-axis and $\hat{\mathbf{y}}_I$-axis, leading to the position error. In contrast, the proposed controller exhibited significantly smaller errors, effectively mitigating these undesirable effects by incorporating rotor dynamics into the control design. \blue{The robust tracking performance under frequently altered rotor directions indicates that our assumption for the negligible reversing delay for aggressive flights is valid.}

For the attitude controller, despite the aggressive rotational trajectory, the attitude error remained small, validating the $\mathbf{R}^T \mathbf{R}_d \approx \mathbf{I}$ approximation. The proposed controller improved the attitude RMSE by 11\%. The relatively modest improvement, compared to significantly impacted positional tracking, can be attributed to the inertia matrix $\mathbf{J}$ being nearly an identity matrix, which rendered the terms $\dot{\boldsymbol{\omega}} \times \mathbf{J}\boldsymbol{\omega} + \boldsymbol{\omega} \times \mathbf{J}\dot{\boldsymbol{\omega}}$ negligible. Furthermore, the complex trajectory triggering simultaneous rotation around all three axes distributed the control efforts, resulting in insufficient angular jerk in any single axis to prominently showcase the effects of rotor dynamics on the control system's performance. 

\section{Conclusion and Future Work} \label{con}

In this paper, we studied the problem of tracking control for omnidirectional multirotors performing aggressive maneuvers, where the rotor dynamics become significant. We designed a novel geometric controller by incorporating a thrust dynamics model for the rotor dynamics, which improves both the positional and rotational control of these systems. The zero equilibrium of the tracking error dynamics is shown to be almost globally exponentially stable using Lyapunov's direct method. Additionally, we validated the proposed controller's stability and tracking performance in three experiments that highlight different aspects of rotor dynamics' impact on the tracking control.

For future work, 
\blue{since the current design does not account for rotor saturation, it could be addressed via constrained least squares.} 
\blue{While reversing delay is not considered in this work, its modeling is an interesting direction to explore, especially for bidirectional rotors.}
Additionally, integrating robust control mechanisms that adjust to changes in rotor characteristics and operational demands might improve the efficacy of the control system.



\balance

\end{document}